\documentclass[11pt,a4paper]{article}
\usepackage[hyperref]{emnlp2020}
\usepackage{times}
\usepackage{latexsym}

\usepackage{microtype}

\usepackage{multirow}
\usepackage{color}
\usepackage{amsmath,amssymb}
\usepackage{mathtools}
\usepackage{graphicx}
\usepackage{wrapfig}
\usepackage{lipsum}
\usepackage{natbib}
\usepackage{caption}
\usepackage{subcaption}
\usepackage{xr-hyper}

\aclfinalcopy 
\def\aclpaperid{2410} 

\title{Human-centric dialog training via offline reinforcement learning}

\author{Natasha Jaques*$^{12}$, Judy Hanwen Shen*$^1$, Asma Ghandeharioun$^1$, Craig Ferguson$^1$, \\
\textbf{Agata Lapedriza$^{1}$, Noah Jones$^1$, Shixiang Shane Gu$^2$, Rosalind Picard$^1$} \\
\\
*Equal contribution \\
  $^1$Massachusetts Institute of Technology, Cambridge, USA \\
  \texttt{$<$judyshen, asma\_gh, agata, ncjones, roz$>$@mit.edu} \\
  $^2$Google Research, Mountain View, USA \\
  \texttt{$<$natashajaques, shanegu$>$@google.com}
  }

\date{}

\begin{document}
\maketitle
\begin{abstract}
How can we train a dialog model to produce better conversations by learning from human feedback, without the risk of humans teaching it harmful chat behaviors? We start by hosting models online, and gather human feedback from real-time, open-ended conversations, which we then use to train and improve the models using offline reinforcement learning (RL). We identify implicit conversational cues including language similarity, elicitation of laughter, sentiment, and more, which indicate positive human feedback, and embed these in multiple reward functions.  A well-known challenge is that learning an RL policy in an offline setting usually fails due to the lack of ability to explore and the tendency to make over-optimistic estimates of future reward. These problems become even harder when using RL for language models, which can easily have a 20,000 action vocabulary and many possible reward functions.  We solve the challenge by developing a novel class of offline RL algorithms. These algorithms use KL-control to penalize divergence from a pre-trained prior language model, and use a new strategy to make the algorithm pessimistic, instead of optimistic, in the face of uncertainty.  We test the resulting dialog model with ratings from 80 users in an open-domain setting and find it achieves significant improvements over existing deep offline RL approaches. The novel offline RL method is viable for improving any existing generative dialog model using a static dataset of human feedback.
\end{abstract}

\section{Introduction}
Training open-domain dialog models is inherently difficult, since for each utterance there are many acceptable responses, yet no perfect response. While supervised learning from conversational corpora allows models to learn grammatical structure and even topic coherence, these models do not generalize, since the training objectives mostly lead the models to memorize responses within the corpus.

\begin{figure}[t]
  \begin{center}
   \includegraphics[width=\columnwidth]{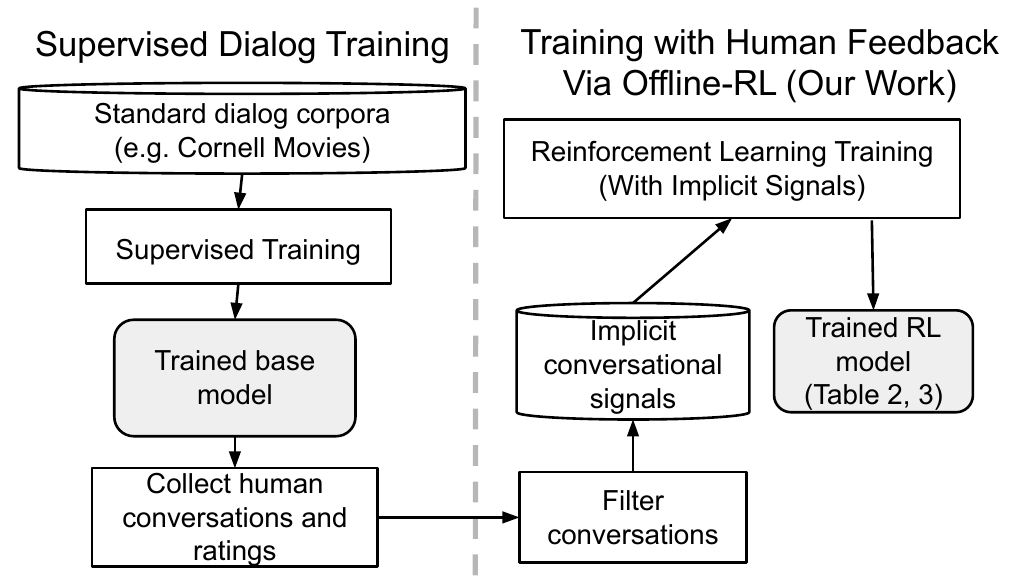}
  \caption{\small Schematic diagram of our method for training with human conversation cues via offline RL. Unlike traditional approaches which stop at using explicit feedback to evaluate static conversations, we allow humans to freely interact with dialog models, and compute metrics based on their implicit satisfaction which are optimized using offline RL. }
  \label{fig:schematic}
  \end{center}
\end{figure}

Humans are the ultimate authority in evaluating what makes one conversational reply better than another. To learn from real conversations with humans, we created an interactive, online platform which hosted a diverse set of neural network dialog models that users could chat with in real time. However, when learning from human interactions in the wild it is crucial to be able to learn offline and test the policy before deploying it, lest it learn inappropriate behaviors (e.g. \citet{tay}). Thus, we need to train and test models offline, to ensure safe model outputs. In order to safely learn to optimize human feedback we pursued an \textit{offline reinforcement learning} approach to training dialog models (see Figure \ref{fig:schematic}). 

Offline RL is challenging; most deep RL algorithms fail to learn from data that is not heavily correlated with the current policy \citep{fujimoto2018off}. Even models based on off-policy algorithms like $Q$-learning fail to learn in the offline RL setting, as the model is not able to explore. If the offline dataset is not sufficient to cover the input-response space, offline RL models suffer from \textit{extrapolation error}, learning arbitrarily bad estimates of the value of responses not contained in the data. 

We solve these problems by developing a new method for offline RL.  The method starts by leveraging a pre-trained language model to constrain offline RL updates. While training with RL, we penalize divergence from this prior model using forms of KL-control. This combats extrapolation error, and ensures that the RL model learns a policy that stays close to the distribution of realistic language, while learning to maximize positive human responses using the offline data. Further, we use dropout to obtain uncertainty estimates of the target $Q$-values, and to obtain a lower bound to alleviate over-optimistic bias in estimating future reward. We show that this new method is able to learn successfully from many different reward functions, even in a very large space with 20,000 tokens.

Both linguistic theory (e.g. Grice's Maxims \cite{grice1975logic}) and empirical experiments correlating human judgement with language features suggest that there are many criteria that could be used to evaluate a conversational agent  \cite{ghandeharioun2019approximating, adiwardana2020towards}. We develop a set of reward functions for our dialog agents to optimize, which are designed to approximate implicit human preferences expressed during conversational responses. We show that the new method is better able to optimize these rewards using the offline data, and when tested with a new set of 80 human conversation partners, leads to more positive responses and higher quality ratings than a state-of-the-art offline deep RL method.

Novel contributions of this paper are:
\begin{itemize}
    \item A new offline RL method, Way Off-Policy (WOP) learning, which introduces the use of KL-control from a pre-trained model to reduce extrapolation error, and an approach to make estimates more pessimistic in the face of uncertainty.
    \item Experiments showing the effectiveness of WOP above strong offline RL baselines.
    \item An investigation into developing conversation rewards based on how human preferences are implicitly expressed in text. We are the first work to learn from implicit signals in conversation using offline RL.
\end{itemize}

\section{Related Work}
\subsection{Dialog}
Improving dialog systems with RL has largely been restricted to task-oriented dialog systems, which have a limited number of task-specific actions \cite{fatemi2016policy, gavsic2011line, liu2017iterative, liu2018dialogue, su2017sample}. Some of these approaches incorporate human input through explicit, manual feedback \cite{shah2018bootstrapping} or implicit signals (e.g. the user interrupting the system or starting over) \cite{shi2018sentiment}. 

RL in the open-domain dialog setting is less explored  \cite{li2016deep,li2017adversarial,li2018dialogue}. Authors may choose to use a highly restricted action space; for example, using RL to choose which dialog model to invoke \citep{serban2017deep}.
\citet{ziegler2019fine} used explicit human feedback to improve the summarization and text continuation performance of a large-scale language model. 

Although implicit signals such as sentiment \cite{hancock2019learning} and conversation length \cite{zhou2018design} have been used in maximum likelihood estimation (MLE) systems, the idea of using such signals as a reward for RL is relatively unexplored. \citet{henderson2008hybrid} combine using reinforcement learning to optimize dialog reward with using supervised learning to restrict the conversation to be close to the training data. \citet{shin2019happybot} use on-policy learning in conjunction with a user-sentiment approximator to improve a seq2seq model, but are unable to learn directly from user feedback.  To the best of our knowledge, we are the first to use offline RL to train dialog models on real human interactions.

\subsection{Offline RL and KL-Control}
The approach we propose is based on KL-control, a branch of stochastic optimal control (SOC)~\cite{stengel1986stochastic} where the Kullback-Leibler (KL) divergence from some distribution is used to regularize an RL policy \cite{abdolmaleki2018maximum, kappen2012optimal,rawlik2012stochastic, todorov2006linearly}. Well-known examples include Trust Region Policy Optimization (TRPO) \cite{slmja-trpo-15}, and use conservative, KL-regularized policy updates to restrict the RL algorithm to stay close to its own prior policy \cite{haarnoja2018soft,kakade2002natural,peters2010relative,rawlik2012stochastic}. KL-control has been used to improve transfer learning between maximum likelihood estimation (MLE) training on data, and training with RL \cite{jaques2017sequence}. Our work is the first to propose KL-control from a pre-trained model to improve offline RL. 

Other strategies to improve off-policy learning differ from our work:  They either have focused on scenarios where the policy is able to explore and collect more data \cite{degris2012off,riedmiller2005neural} such as learning online from an outdated replay buffer (e.g. \cite{munos2016safe}), or have performed off-policy policy evaluation \cite{farajtabar2018more,jiang2016doubly,precup2000eligibility,thomas2016data}. In contrast, we learn a policy entirely offline, from a fixed batch of data, with no ability to explore. Others have tackled this problem using deep learning, but have not used KL-control
~\cite{liu2019off,gelada2019off,bhatt2019crossnorm,kumar2019stabilizing,agarwal2019striving,fujimoto2018off, ghasemipour2020emaq}.

Most similar to our work is Batch Constrained Q-learning (BCQ) \cite{fujimoto2018off}, which addresses extrapolation error in offline RL by  constraining the actions of the policy to be close to the offline data. This is accomplished by learning a generative model of the offline data, $p(a|s)$, and sampling from this model during learning and inference. We improve upon this approach by using KL-control to directly integrate knowledge of the prior model $p(a|s)$ into the RL policy. 

\section{Way Off-Policy RL}
\label{sec:methods}
\begin{table*}[t]
\resizebox{\textwidth}{!}{%
\begin{tabular}{l|l}
\begin{tabular}[c]{@{}l@{}}
    \textbf{{[}User{]}}: hey, what's up? \\ 
    \textbf{{[}Batch Q{]}}: where did you get??, that sounds so exciting. how \\do you work for you... you want to talk to me?... you want\\
    \textbf{{[}User{]}}: yes, i want to talk to you! \\ 
    \textbf{{[}Batch Q{]}}: where did you say to me?... you want to talk to \\ you about you... you are so good for you... you are so good...\\
\end{tabular} & 
\begin{tabular}[c]{@{}l@{}}
    \textbf{{[}User{]}}: hey, what's up? \\ 
    \textbf{{[}KL-control{]}}: hey, i hope you have a great day and \\ i wish you the best. \\
    \textbf{{[}User{]}}: oh thank you that's very sweet of you. \\ 
    \textbf{{[}KL-control{]}}: so, i'm so excited, and i'm so excited \\ to meet new people.
\end{tabular} 
\end{tabular}
}
\caption{\small Purely reward-maximizing methods like \textit{Batch $Q$} trivially exploit a reward for asking questions by only asking questions, and using the maximum number of tokens in every sentence. In contrast, KL-control methods output plausible language by staying close to the language prior, while eliciting positive feedback from humans.}
\label{tab:samples}
\end{table*}
We adapt typical RL notation to the problem of generating a conversation. Here, we consider human interaction to represent the RL environment. The conversation history is the state $s_t$ of the environment at timestep $t$, and is composed of a series of utterances, which are composed of vocabulary tokens. The action $a_t$ that the RL model must take at each timestep is to select the most appropriate token according to its policy $\pi(a_t|s_t)$. Once it has constructed an utterance, the response of a human to that utterance is used to compute a reward signal $r_t$ to train the model.
The agent's goal is to maximize reward over a conversation trajectory $\tau$, with a discount factor of $\gamma$ applied to future rewards.  

$Q$-learning methods learn an action-value estimate of the total expected discounted future reward, $Q_{\pi}(a_t,s_t) = \mathbb{E}_{\pi}[\sum_{t'=t}^{T} \gamma^{t'-t}r_{t'}]$, through iterative updates based on the Bellman equation:
\begin{align} 
\label{eq:bellman}
\begin{split}
    {}& Q_{\theta_\pi}(s_t,a_t) =  r_t +  \\
    &\gamma \mathbb{E}_{s_{t+1}\sim p(\cdot|s_t,a_t)}[\max_{a_{t+1}} Q_{\theta_T}(s_{t+1},a_{t+1})] 
\end{split}
\end{align}
In deep $Q$-learning \citep{dqn}, a $Q$-network approximates $Q_{\theta_\pi}(s_t,a_t)$ and drives the policy $\pi$. A second Target $Q$-network approximates the expected reward from the next state, $Q_{\theta_T}(s_{t+1},a_{t+1})$ \citep{van2016deep}. Here, we used pre-trained language models to initialize our $Q$- and Target $Q$- networks.

\subsection{Offline RL and extrapolation error}
In offline RL, we are given a fixed batch of data $\mathcal{B}$, and assume that no further interaction with the environment is possible. To train $Q_{\theta_{\pi}}$, we sample $(s_t, a_t, r_t, s_{t+1}) \sim \mathcal{B}$, and update the weights of the $Q$-network to approximate Eq. \ref{eq:bellman}. Because $Q$-learning is an off-policy algorithm, in principle it should be able to learn from data collected by any behavior policy. However, extrapolation error occurs when the ORL policy learns to favor a state-action pair $(a,s)$ that is unlikely, or not contained, in the batch data. In this case, the estimate $Q(a,s)$ can be arbitrarily bad \citep{fujimoto2018off}. Because the Bellman equation bootstraps each $Q$-value based on all future $Q$ estimates, any error can accumulate to distort $Q$-values \cite{kumar2019stabilizing}. Experiments from \citet{fujimoto2018off} show that extrapolation error can be highly detrimental to offline RL.

These problems are compounded by the fact that algorithms like $Q$-learning are inherently optimistic in the face of uncertainty. When value estimates for some region of the state-action space are noisy (because too few experience samples have been used to refine them), the maximum operation in Eq. \ref{eq:bellman} will lead to an overestimation of expected reward. In a normal RL setting, this overestimation bias drives the model to explore states and actions for which the value estimates have the highest variance, thus enabling it to refine them; in essence, creating a built-in drive to explore. In the offline setting, where exploration is not possible, the model is instead driven to value parts of the state-action space for which it has little to no data to learn a good policy. 
Table \ref{tab:samples} shows an example of this effect, where a vanilla $Q$-learning model trained on an offline batch of data (\textit{Batch $Q$}) begins to use unrealistic language that is not contained within the batch data, for example saying implausible phrases such as ``\textit{where did you say to me?}". 

Even in the online setting, applying deep RL to dialog generation is challenging due to the large state-action space. While typical game RL tasks may have an action space of dimension 8 \cite{dqn}, in dialog the action space is the number of tokens in the vocabulary: 20,000. The high-dimensional state-action space further compounds the problems of extrapolation error and overestimation bias in offline RL. Below, we describe a novel method to ameliorate these issues.

\subsection{Dropout for uncertainty estimation of Target $Q$-values}
\label{sec:dropout}
Overestimation error in estimating future rewards based on Target $Q$-values poses an issue for offline RL. We leverage the fact that a network trained with dropout can be used to approximate a Bayesian uncertainty estimate of the network's output \citep{gal2016dropout}. 
Given the target $Q$-network $Q_{\theta_T}$, we compute $Q(a_{t+1}, s_{t+1})$ by running $M$ stochastic forward passes of the network, each with a new dropout mask $d_i$. Taking the minimum of these outputs gives a Monte Carlo (MC) estimate of the lower-bound of $Q_{\theta_T}(a_{t+1}, s_{t+1})$:
$$Q(a_{t+1}, s_{t+1}) = \min_{i=1...M}[Q_{\theta_T}(a_{t+1}, s_{t+1}; d_i)]$$
This penalizes high variance estimates and leads the algorithm to be pessimistic in the face of uncertainty, rather than optimistic, favoring actions and states well covered by the offline data. 

\subsection{KL Control from pre-trained prior}
\label{sec:klcontrol}
Recall that BCQ \cite{fujimoto2018off} uses offline data to learn a model of which actions are probable given a state: $p(a|s)$. It then samples actions from $p(a|s)$ to constrain the RL policy such that it cannot take unrealistic actions. 

In the language domain, we already have access to a better model of $p(a|s)$ than could easily be learned from a small amount of offline data. Any language model gives us the probability of a word occurring given a particular conversation context ($p(a|s)$), and can be used as a language \textit{prior} to prevent the RL model from choosing unrealistic words. Rather than simply sampling from this prior, we directly incorporate knowledge of the prior into the RL policy. To achieve this, we use KL-control to penalize divergence between the prior $p(a|s)$ and the $Q$-network policy $\pi_{\theta}$, while maximizing reward.

Given a trajectory of actions, $\tau=\{a_1, a_2, ... a_{t-1}\}$, let $q(\tau) = \prod_{t=1}^T \pi_{\theta}(a_t, s_t)$ be the policy of our $Q$-learning algorithm at the trajectory level. Similarly, let $p(\tau) = \prod_{t=1}^T p(a_t|s_t)$ be the prior distribution over the trajectory, and $r(\tau)$ be the rewards. We seek to maximize the following KL-regularized objective:
\begin{align}
    L(q) = \mathbb{E}_{q(\tau)}[r(\tau)]/c - D_{KL}[q(\tau)||p(\tau)]
\end{align}

As $D_{KL}[q||p] = \sum_x q(x)(\log q(x)-\log p(x))$, this is equivalent to maximizing the following expected value function at the action level:
\begin{align}
    \vspace{-0.5cm}
    \label{eq:kl_value}
    \begin{split}
    Q^{\pi}(s_t, a_t)& = \mathbb{E}_{\pi}[\sum_{t'=t}^T r(s_{t'}, a_{t'})/c \\
    &+ \log p(a_{t'}|s_{t'}) - \log \pi(a_{t'}|s_{t'})]
    \end{split}
    \vspace{-0.5cm}
\end{align}
The two terms we have introduced in Eq. \ref{eq:kl_value} have clear implications. The $\log p(a|s)$ term rewards choosing actions that have high probability under the prior, biasing 
the model to state-action pairs that are realistic and likely to be in the offline data; thus, extrapolation error is reduced. The effects of using KL-control to ensure an RL model continues to use realistic language are shown in Table \ref{tab:samples}.

The $-\log \pi(a|s)$ term is analogous to entropy regularization. Maintaining diversity through entropy regularization is important for dialog models, which are known to collapse to a small number of uninteresting samples \cite{li2016dialogue}. 

We can derive an entropy-regularized version of $Q$-learning, known as soft $Q$-learning \cite{haarnoja2017reinforcement}, or $\Psi$-learning \cite{jaques2017sequence,rawlik2012stochastic}. This allows us to re-state our entropy-regularized, KL-control objective as:
 
\begin{align}
    \begin{split}
    \Psi^*(s_t, a_t) &= r(s_{t'}, a_{t'})/c + \log p(a_{t'}|s_{t'})  \\
    &+ \gamma \log \sum_{a'}\exp(\Psi^*(s', a'))
    \end{split}\\
    \pi^*_{\Psi}(a_t|s_t) &= \exp(\Psi^*(s_t, a_t))
\end{align}
Because it avoids taking a hard max over noisy estimates, this $\Psi$-learning objective leads to less overestimation of future reward, and aids learning through more stable temporal-difference updates. 

\subsection{Comparison to existing techniques}
To test our algorithm against a state-of-the-art offline deep RL technique, we implement a discrete version of Batch Constrained Q-learning \cite{fujimoto2018off}, DBCQ. For a fair comparison, we also use a fully trained language model to provide $p(a|s)$ to BCQ, and apply our Monte Carlo target estimation technique to reduce overestimation error. Finally, to adapt BCQ to discrete action spaces, we remove the continuous-action perturbation model. 

\section{Learning from talking to humans}
Figure \ref{fig:schematic} illustrates our experimental approach. The left side of the figure describes traditional approaches to dialog generation, in which human feedback is only used to evaluate static  conversations generated by dialog models. In contrast, we allow humans to freely interact with our models online, and use their implicit conversation cues to update our dialog models using offline RL. 

\subsection{Training baseline dialog models}
\label{sec:vanilla_training}
Before learning from human feedback with RL, we first train a collection of baseline dialog models using standard corpora: the \textsc{Cornell} dataset of movie dialog \cite{cornell_dataset} and a \textsc{Reddit} Casual Conversations dataset \cite{ghandeharioun2019approximating}. For model architectures, we focused on hierarchical sequence-to-sequence models \cite{serban2016building,serban2017hierarchical,park2018hierarchical}

because they were found to be more effective for the datasets under consideration than e.g. Transformers \cite{saleh2019hierarchical}. Regardless, the techniques proposed here are model-agnostic, and could be applied to a dialog model with any underlying architecture. In total, we trained over 40 dialog models with different architectures, on different datasets, with different feature-based regularization (e.g. sentiment or relatedness as in \citet{ghandeharioun2019approximating}). These models vary significantly in the distribution of language they learned, and thus differ significantly from the offline RL policy. 

\subsection{Hosting real-time conversations online}
\begin{figure*}[h]
  \begin{center}
   \includegraphics[width=\textwidth]{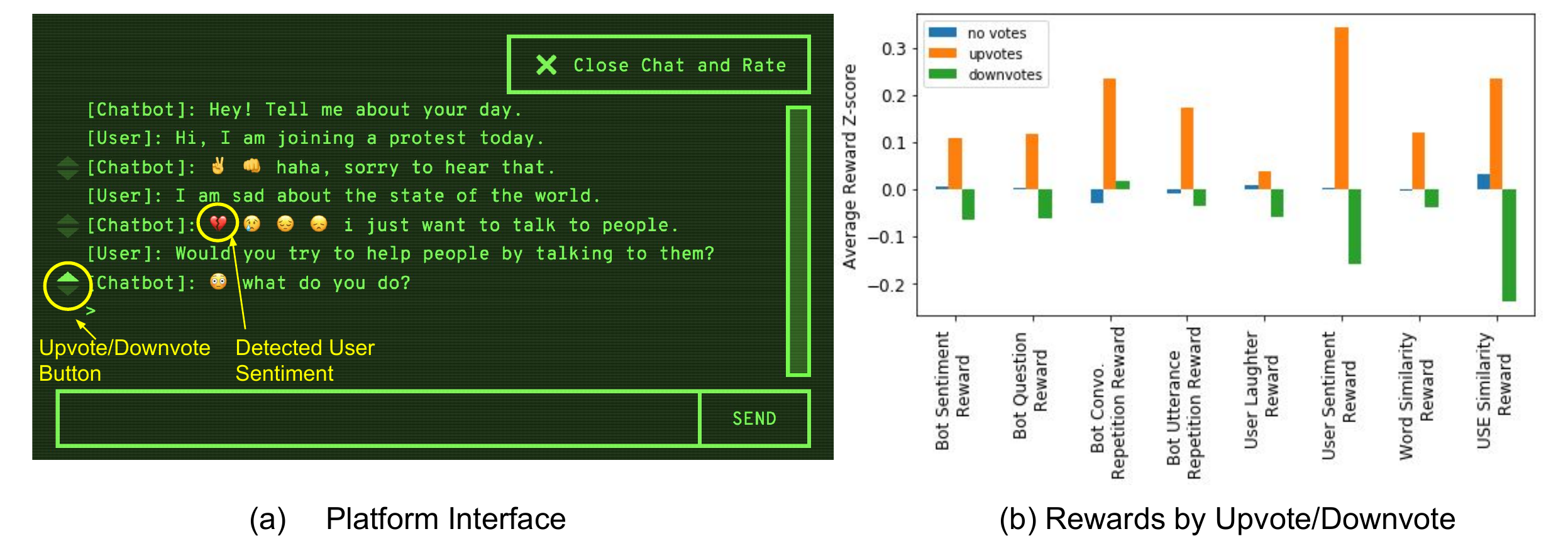}
  \caption{\small (a) Platform interface in which users chat in real time with dialog models hosted on GPU. The interface displays the user's sentiment detected with DeepMoji \cite{felbo2017using}, and includes buttons for the user to upvote (downvote) a response they particularly like (dislike). (b) By conditioning on responses which received positive, neutral, and negative manual feedback (votes), we can determine which implicit rewards map most clearly to user ratings.}

  \label{fig:interface}
  \end{center}
\end{figure*} 

The trained models were deployed to interact live with human users via a web server that hosts neural network dialog models on GPU for fast, real-time inference: \url{https://github.com/asmadotgh/neural_chat_web}. Figure \ref{fig:interface} shows a screenshot of the interface, which includes buttons that allow users to give manual feedback on responses they particularly liked or disliked. Users were encouraged to use these buttons, and we sum these manual votes to create an overall \textit{votes} score.
After chatting, users were asked to provide a Likert scale rating of the bot's conversation quality, fluency, diversity, contingency/relatedness, and empathy. The code for the RL models is available in open-source at \url{https://github.com/natashamjaques/neural_chat/tree/master/BatchRL}. Using the server, we collected a batch of human interaction data containing $46,061$ pairs of user input and agent response. Because humans may use inappropriate language with bots online (see \cite{tay}), 
we filtered this data to remove 1 character responses, profanities, and invalid inputs for a remaining total of $45,179$ response pairs. This filtering step is important to ensure undesirable human behavior is not learned by the RL algorithms. The offline data was used to train the RL models as described in Section \ref{sec:methods}.

\subsection{Evaluating offline RL models}
We recruited 80 Mechanical Turk workers to provide a total of 600 7-point Likert scale ratings of the trained bots, after interacting with each for at least $6$ turns. We note that using this platform to test our models ``in the wild" with novel humans represents a more meaningful test of generalization than testing an RL model in the same limited (game) environment in which it was trained, since humans are not restricted in the text they can type as input to the model.

\section{Measuring implicit conversation cues}
\label{sec:implicit}

Our goal is to improve a dialog model's ability to engage in natural conversation with a human by learning from the implicit signals in the human's response. Requiring a human to manually rate good interactions is unnatural and cumbersome, and we hypothesize it cannot scale as effectively as recognizing and learning from informative cues within the user's text responses. The golden question is which goals should be used to train a good chit-chat dialog model.

Understanding when a human is satisfied with the conversation is an unsolved problem. As a first step, 
we designed several intrinsic conversation rewards, taking inspiration from prior work in dialog, as well as the psychology of human conversation. We noted that psychologists have identified the importance of emotion in creating a sense of understanding \cite{bodie2015role, weger2010active}, laughter as important to building solidarity \cite{hay2000functions}, paraphrasing and style matching as helping to facilitate good conversation \cite{ireland2011language, weger2010active}, and asking questions as an important active listening skill \cite{bodie2012listening}. Further, prior work has found that eliciting longer conversations can be a signal of engagement \cite{sidner2004look,zhou2018design}, and that reducing repetition and increasing specificity on the part of the model can improve conversation quality \cite{see2019makes, mehri2020unsupervised}. We compute a large collection (30 in total) of bot rewards (rewards based on bot behavior e.g. asking questions), user rewards (rewards based on eliciting positive user behavior e.g. laughter), and interaction rewards (rewards based on similarity between the user's input and bot's response e.g. similarity to the user's response in sentence embedding space). 

\begin{figure*}[ht!]
  \begin{center}
   \includegraphics[width=\textwidth]{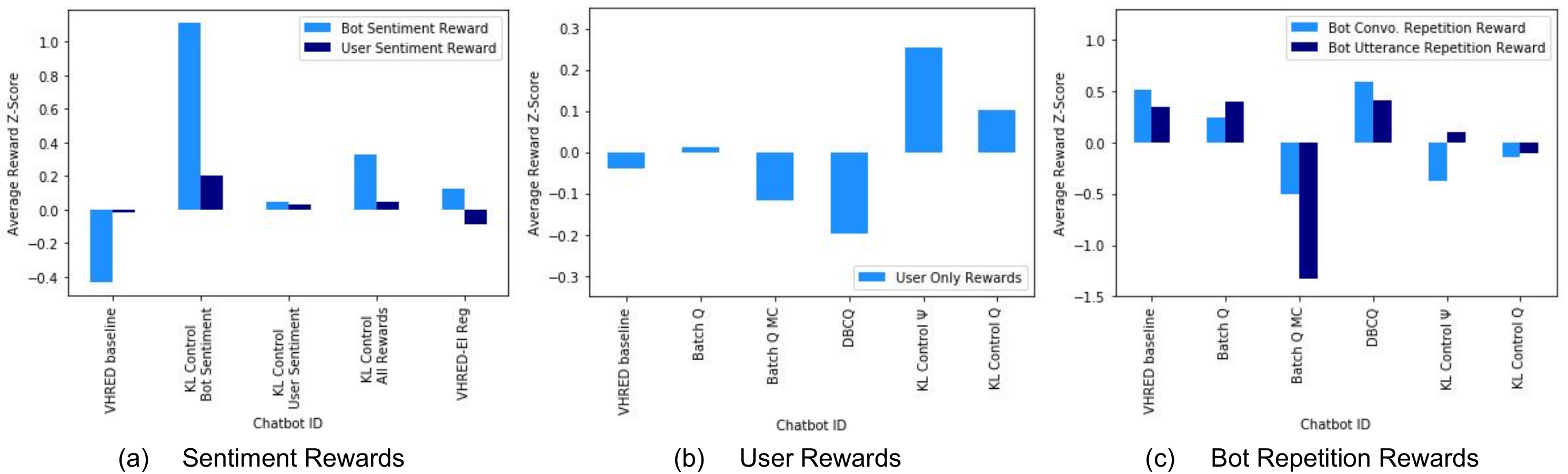}
  \caption{\small (a) Average reward scores of sentiment rewards computed on study chat transcripts across different models. KL-control methods more effectively increase bot sentiment and elicit more positive sentiment from humans than either the baseline language model or adding sentiment regularizer during supervised training. (b) The sentiment and laughter elicited from humans is higher for KL-control methods than the language model baseline and other offline RL techniques. (c) Average bot repetition reward scores (higher scores indicate less repetition). The RL models contain more conversation and utterance repetition.}
  \label{fig:results}
  \end{center}
\end{figure*} 

To determine which of these rewards objectively relate to user satisfaction, we examine the reward score for those responses that received positive, negative, and neutral manual feedback using the upvote/downvote buttons provided in the interface. We found that only some of the rewards mapped accurately to user ratings (see Figure \ref{fig:interface}b), and these are the ones we optimize with our RL models. For more details about the reward functions, please see the appendix. Notably, conversation length and specificity score were not found to be higher in upvoted bot responses. 

Note that four of the rewards (starting with the \textit{bot} prefix) can be optimized by the model itself, but the remaining four rewards include eliciting positive responses from a human user or measuring user-bot response similarity (e.g. using word overlap or similarity in Universal Sentence Encoder (USE) embeddings \cite{cer2018universal}). 

\section{Results}
\label{sec:results}
\subsection{Controlling bot conversation behavior}
We first examine whether our algorithms can successfully maximize the proposed bot rewards as intended\footnote{In the appendix, we provide a study comparing WOP to prior work in traditional, non-dialog RL tasks, and find that it outperforms all relevant baselines including DBCQ.}. We trained RL models on 1) bot sentiment reward only, 2) user sentiment reward only, and 3) a combination of rewards (from Figure \ref{fig:interface}b). We compare the effectiveness of these models to a baseline VHRED model and a Sentiment and Infersent regularized VHRED model (as proposed by \citet{ghandeharioun2019approximating}). We compute the reward scores (e.g. sentiment) based on conversations with new humans in the wild (i.e. during the final study). Figure \ref{fig:results}a shows that the KL-control model, trained to maximize bot sentiment, achieves higher bot sentiment in experiments than both the VHRED baseline and the VHRED-EI model (with sentiment and topic regularization \cite{ghandeharioun2019approximating}). This illustrates that for controlling bot sentiment, a reward-based approach better optimizes bot behavior than training with sentiment-based regularization. Furthermore, controlling bot sentiment also leads to eliciting higher user sentiment in our open-domain experiments.

\subsection{Measuring human conversation behavior}
We then consider how effective our algorithms are at maximizing rewards that are based on human behavior. 

Although user rewards are inherently more difficult to optimize than bot rewards, Figure \ref{fig:results}b illustrates that our KL-control models elicit higher human reward scores (user sentiment and user laughter) than other offline RL algorithms and the baseline VHRED model. This demonstrates the success of our algorithms in eliciting positive responses from the human conversation participants\footnote{In the appendix, we replicate these experiments with a different baseline model, and produce the same findings.}. 

\subsection{Overall human ratings}
Table \ref{tab:techniques_human} shows the results of the human evaluation, comparing WOP to ablations of itself, vanilla offline RL (Batch $Q$), and DBCQ. 

Compared to the RL baseline (Batch $Q$), MC Target $Q$ estimation leads to modest improvements in Fluency. While the DBCQ model is rated better than Batch $Q$ and does well in the Diversity category, it performs worse than the WOP KL-control methods, particularly at eliciting human rewards. The KL-control models show substantial gains over the RL baselines across both ratings and human reward. We perform a one-way analysis of variance (ANOVA) comparing the KL-control models to the Batch $Q$ baselines and DBCQ on total human ratings, and find that the KL-control models are significantly better, $F(x) =  7.328, p <.005$. This validates the hypothesis that KL-control with a strong, pre-trained prior can be used to improve offline RL. 

\subsection{The role of repetition}
The overall human quality ratings are worse in the offline RL bots as compared to the language model prior (Table \ref{tab:techniques_human}). The biggest gap between the VHRED and RL models is the diversity ratings. The conversation and utterance repetition scores of each technique in Figure \ref{fig:results}c reveal that the RL models (including the KL-control models) contain more repetition than the baseline. We hypothesize that due to the limited size of our offline data, the RL models have restricted their outputs to focus on a narrow range of conversations that elicited high rewards in the training data, which may increase repetitiveness. Some applications may require shaping dialog model behavior towards a desired objective (such as using appropriate language) over maximizing other conversation objectives.

\begin{table*}[t]
\resizebox{\textwidth}{!}{
\begin{tabular}{|l|lllll|l|r|r|}
\hline
\textbf{Model type} & \textbf{Quality} & \textbf{Fluency} & \textbf{Diversity} & \textbf{Relatedness} & \textbf{Empathy} & \textbf{Total}  & \textbf{Votes} &
\textbf{\begin{tabular}[c]{@{}l@{}}Human\\ reward\end{tabular}} \\ \hline
VHRED-Baseline  & 2.65 $\pm$.46          & 3.83 $\pm$.47          & 4.05$\pm$.52            & 2.43 $\pm$.44          & 3.08 $\pm$.53            & 16.03 $\pm$1.93           & 0.27        & -0.04  \\ \hline
DBCQ            & 1.80  $\pm$.41         & 1.49  $\pm$.29         & \textbf{3.22 $\pm$.57}  & 1.56 $\pm$.25          & 2.10  $\pm$.37           & 10.17 $\pm$1.29           & -0.07        & -0.20    \\
Batch $Q$         & 1.30  $\pm$.19         & 2.85 $\pm$.54          & 1.15 $\pm$.13           & 1.23 $\pm$.15          & 2.18 $\pm$.55            & 8.70 $\pm$0.97            & -0.16        & 0.01    \\
Batch $Q$ + MC    & 1.53 $\pm$.24          & 2.15 $\pm$.37          & 1.60 $\pm$.32           & 1.53 $\pm$.28          & \textbf{2.58 $\pm$.48}   & 9.38 $\pm$1.31            & -0.21        & -0.12    \\
KL-control $Q$    & \textbf{2.23 $\pm$.44} & \textbf{2.88 $\pm$.41} & 2.65 $\pm$.41           & \textbf{2.15 $\pm$.39} & 2.28 $\pm$.47            & \textbf{12.18 $\pm$1.59}  & \textbf{0.09} & 0.10     \\
KL-control $\Psi$ & 1.98 $\pm$.44  & 2.73 $\pm$.45          & 2.30 $\pm$.42           & 1.90 $\pm$.37          & 2.40 $\pm$.44            & 11.30 $\pm$1.63           & 0.04          & \textbf{0.25} \\ \hline

\end{tabular}
}
\caption{\small Interactive human evaluation of offline RL techniques (best RL model bolded). KL-control strongly outperforms other offline RL techniques. Ratings are Likert scale with 95\% confidence intervals ($n=40$). Votes and human reward are $z$-scores.}
\label{tab:techniques_human}
\end{table*}

\begin{table*}[t]
\centering
\resizebox{\textwidth}{!}{%
\begin{tabular}{|l|lllll|l|r|r|}
\hline
\textbf{\begin{tabular}[c]{@{}l@{}}Reward\\ function\end{tabular}} & \textbf{Quality} & \textbf{Fluency} & \textbf{Diversity} & \textbf{Relatedness} & \textbf{Empathy} & \textbf{Total}  & \textbf{Votes} & \textbf{\begin{tabular}[c]{@{}l@{}}Human\\ reward\end{tabular}}  \\ \hline
Manual votes    & 2.53 $\pm$.51          & 3.43 $\pm$.52          & 2.88 $\pm$.50          & 2.40 $\pm$.45          & 3.30 $\pm$.45            & 14.53 $\pm$1.96          & -0.05        & -0.07          \\
User laughter   & 2.53 $\pm$.47          & 3.38 $\pm$.50          & \textbf{3.05 $\pm$.47} & 2.25 $\pm$.43          & 3.08 $\pm$.48            & 14.28 $\pm$1.96          & 0.06        & 0.01          \\
User Sentiment  & \textbf{2.60 $\pm$.49} & 3.30 $\pm$.50          & 2.90 $\pm$.50          & 2.38 $\pm$.47          & 3.23 $\pm$.55            & 14.40 $\pm$2.25          & 0.04        & 0.05          \\
Word Similarity & 2.58 $\pm$.52          & 3.53 $\pm$.49          & 2.98 $\pm$.50          & 2.45 $\pm$.45          & 3.08 $\pm$.46            & 14.60 $\pm$2.00          & 0.02        & -0.18          \\
USE Similarity  & 2.05 $\pm$.41          & 3.65 $\pm$.48          & 2.38 $\pm$.46          & 2.03 $\pm$.45          & 2.75 $\pm$.46            & 12.85 $\pm$1.77          & -0.11        & -0.11          \\
Bot Question    & 2.43 $\pm$.52          & 3.65 $\pm$.52          & 2.63 $\pm$.47          & 2.65 $\pm$.51          & 2.70 $\pm$.48            & 14.05 $\pm$2.14          & 0.01        & 0.09          \\
Bot Sentiment   & 1.90 $\pm$.45          & 3.20 $\pm$.53          & 1.88 $\pm$.52          & 1.88 $\pm$.46          & 3.20 $\pm$.41            & 12.05 $\pm$1.91          & -0.04        & \textbf{0.14}  \\
Bot Repetition  & 2.48 $\pm$.45          & \textbf{3.78 $\pm$.49} & 2.95 $\pm$.52          & \textbf{2.63 $\pm$.45} & \textbf{3.65 $\pm$.61}   & \textbf{15.48 $\pm$1.97} & \textbf{0.07} & 0.05 \\ \hline
\end{tabular}
}
\caption{\small Interactive human evaluation of WOP trained with different reward functions. Manual votes are outperformed by implicit signals. Ratings are Likert scale with 95\% confidence intervals ($n=40$), votes and human reward are $z$-scores.
}
\label{tab:rewards_human}
\end{table*}

\subsection{Comparing rewards}
Table \ref{tab:rewards_human} presents the results of models trained with only a single reward function, to investigate which rewards presented in Section \ref{sec:implicit} are useful for achieving high-quality conversations with humans. 

We note that extracting a set of reward functions post-hoc from a batch of data and training on these independently is made feasible through offline RL. 
Here all models are trained with WOP (KL-control, $\Psi$-learning, and MC targets). Maximizing positive sentiment in the user leads to the highest quality bot, underscoring the importance of implicit signals as cues for good conversation. The bot trained on the manual votes provided by users at the utterance level achieves decent quality scores, but fails to elicit a higher z-score of manual upvotes than other models. 

Training on the manual upvote reward may help the bot learn successful behaviors indirectly but such a sparse reward is difficult to optimize for directly. Even though users were instructed to make use of the vote feature, voting is burdensome, and users did not vote frequently enough to provide a good training signal. 
 
Meanwhile, \textit{implicit} signals of human enjoyment (such as sentiment) are dense and thus a more scalable way to learn from human preferences. Across all bots trained on single features, the bot trained on minimizing repetition (both on a conversational and utterance level) achieves the best quality over all. 

\section{Discussion}
In this work, we present novel techniques that enable successful offline reinforcement learning on any base language model from real human conversations. This allows the dialog systems practitioner to train models that learn language structure from vast, readily-available corpora, then fine-tune for specific desirable behaviors post-hoc through RL rewards. 

We observe that the new offline RL method successfully optimizes both generated bot rewards and elicited human responses. We show that it presents a better option than using regularization in training a specific bot behavior. Further, RL currently remains the only option for maximizing user feedback over the course of a conversation. 

Compared to prior work in offline RL, the novel WOP offline RL algorithm achieves higher performance in traditional RL tasks, elicits more positive feedback in conversations with novel humans at test time, and earns overall higher human ratings.

A limitation of our study is that the question of what to optimize with RL to improve overall qualitative ratings remains open.  We have shown that manual ratings are too sparse to optimize effectively, and instead suggest using implicit rewards. However, our reward set proved insufficient to achieve higher human quality ratings, at least with the limited offline training data we were able to collect. It is unlikely the rewards proposed here fully cover what it means to have a high quality open-ended conversation. Future work should investigate more rewards for training an open-domain dialog model such as long term conversation rewards that may need to be computed over many conversation turns. 

Our work computes conversational rewards based on dialog data and annotations from online task workers in the United States. Considering the broader impacts of our work, a representative and diverse set of conversations and annotations should be collected before real world systems are trained and deployed using our algorithms. 

We have shown that the proposed techniques can be useful for shaping dialog model behavior towards a desired objective. For many practical applications, we may have specific requirements for the language generated by a model---for example, that it is appropriate, positive, and polite---even if this leads to a lower perception of conversation quality for some users. We have shown that the Way Off-Policy algorithm provides a more effective way to teach a language model specific behaviors from offline data than previously proposed RL or regularization techniques.

\section*{Acknowledgments}
We would like to thank Scott Fujimoto for insightful email correspondence on this topic, approval of the DBCQ algorithm, and suggestion to apply model averaging. We would like to thank Sudha Rao and Yonatan Bisk for helpful guidance and feedback in the re-framing and re-writting process of this work. We also thank Max Kleiman-Weiner, Ardavan Saeedi, Sebastian Zepf, Sara Taylor, Oliver Saunders Wilder, Kyle Kastner, Marissa Zhang, and Kristy Johnson for their helpful discussions about this project, and many others for helping test-drive our bots.

We thank the MIT Quest for Intelligence, and MIT Stephen A. Schwarzman College of Computing, and the Machine Learning Across Disciplines Challenge for providing computing resources, and MIT Media Lab Consortium for the support of this research. This work has been partially supported by RTI2018-095232-B-C22 grant from the Spanish Ministry of Science.

\bibliography{emnlp2020}
\bibliographystyle{acl_natbib}
\appendix
\section{Reproducibility}
\subsection{Training details and hyperparameters}
\label{sec:appendix-model-details}
\subsubsection*{Baseline Models}

The underlying architecture of the baseline language models employed for this work is a Variational Hierarchical Recurrent Encoder Decoder (VHRED) \citep{serban2017hierarchical}. We also conduct a second set of experiments on an enhanced version of this model with additional knowledge distillation to improve the model's ability to track the sentiment and semantics of the conversation, as proposed by \citet{ghandeharioun2019approximating}. The language models were originally trained on two datasets: movie dialogs \citep{cornell_dataset} and a dataset scraped from \url{reddit.com/r/casual_conversation} \citep{ghandeharioun2019approximating}.

The underlying parameters of the VHRED model were as follows: Context RNN hidden size $=1000$, decoder hidden size $=1250$, encoder hidden size $=1250$, $z$ embedding size $=600$, gradient clip $=1.0$, dropout $d=0.2$. The maximum conversation length was fixed at 5 utterances (context from more than 5 utterances ago was discarded), and the maximum sentence length was 30 tokens. The VHRED model has $76.6$ million parameters. 

We also added layers to the Context RNN and regularized it to be able to predict the semantic content of the input utterance using a form of knowledge distillation \citep{hinton2015distilling} from a state-of-the-art sentence-embedding model \citep{conneau2017supervised}. There were 2 additional feedforward semantic prediction prediction layers of size 128, which used ReLu activation. The VHRED model with sentiment and infersent regularization has $95.4$ million parameters.
 
Each RL model was trained on a NVIDIA GeForce GTX 1080 GPU. 

 \subsubsection*{RL Models}
The RL models, the main focus of our work, were trained using human conversation data collected via the online interactive platform (described in Section \ref{sec:appendix-interactive}) and batch size was fixed at 32. Each model was trained for $2000$ epochs. The RL models were initialized with the weights of the best model trained on the Reddit dataset. Early stopping was used to determine the number of training iterations of the best checkpoint. For each bot, 3 different stopping epochs were tested and the best was selected. The checkpoint was selected using manual tuning based on interactive chat with the chatbots. For the best performing bots, KL-Control $Q$ and KL-Control $\Psi$, the 1600 and 1800 epoch checkpoints were selected respectively. 

The reward weights were also tuned to determine which weighting of rewards produced the desired bot behavior. We tried uniform weights (summing up to 1) and slightly increased weights for repetition rewards and human bot interaction rewards. The best weights were found to be assigning $0.15$ to repetition and human bot interaction rewards and $0.1$ to all other rewards. Reward weights were also determined using manual tuning and conversational interaction. The same reward weights were shared between all RL models we trained. Only 3 sets of weights were tried in the reward weights hyperparameter optimization process. 

All other hyperparameters were shared between RL models, and were as follows: discount $\gamma=0.5$, weight placed on RL reward vs. KL-divergence term $c=2$, number of Monte Carlo samples of the Target $Q$-network $M=5$, target network update rate $\alpha=.005$, learning rate $r=.0001$. We used a smooth $L1$ loss function to approximate the $Q$-values, and clipped gradients at a value of $1.0$. The RL models have a total of $76.6$ parameters (same as the VHRED models). 

\begin{table*}[t]
\resizebox{\textwidth}{!}{
\begin{tabular}{|l|lllll|l|r|r|}
\hline
\textbf{Model type} & \textbf{Quality} & \textbf{Fluent} & \textbf{Diverse} & \textbf{Related} & \textbf{Empathy} & \textbf{Total}  & \textbf{Votes} & \textbf{\begin{tabular}[c]{@{}l@{}}Human\\ reward\end{tabular}} \\ \hline
VHRED-EI Baseline   & 3.11  $\pm$.41         & 4.34  $\pm$.44         & \textbf{4.66 $\pm$.49}  & 3.02  $\pm$.47          & 3.45  $\pm$.47          & 18.59 $\pm$1.76          & 0.19          & -0.05\\ \hline
DBCQ                & 1.64  $\pm$.48         & 1.87  $\pm$.34         & \textbf{3.13 $\pm$.58}  & 1.84  $\pm$.34          & 2.09  $\pm$.38          & 10.58 $\pm$1.55          & -0.23         & -0.02 \\
Batch $Q$             & 1.87  $\pm$.30         & 2.36 $\pm$.42          & 2.20 $\pm$.41           & 1.91 $\pm$.32           & 2.58 $\pm$.47           & 11.91 $\pm$1.58          & -0.16         & 0.00 \\
Batch $Q$ + MC        & 1.85  $\pm$.39         & 2.46 $\pm$.44          & 2.46 $\pm$.52           & 1.98 $\pm$.39           & 2.34 $\pm$.49           & 11.07 $\pm$1.82          & -0.07         & \textbf{0.03}  \\
KL-control $Q$        & \textbf{2.38 $\pm$.39} & 3.24 $\pm$.47          & 3.42 $\pm$.54           & \textbf{2.38 $\pm$.45}  & 2.56 $\pm$.43           & 13.98 $\pm$1.81          & 0.02          & 0.01  \\
KL-control $\Psi$ (WOP)  & 2.33 $\pm$.41            & \textbf{3.73 $\pm$.53}  & 2.82 $\pm$.50            & 2.31 $\pm$.44  & \textbf{3.47 $\pm$.50}  & \textbf{14.67 $\pm$1.82} & \textbf{0.13} & \textbf{0.03} \\ \hline
\end{tabular}
}
\caption{Interactive human evaluation of offline RL techniques on the VHRED-EI Model. Ratings are Likert scale with 95\% confidence interval ($n=45$), votes and human reward are $z$-scores.}
\label{tab:techniques_human_EI}
\end{table*} 

\begin{table*}[t]
\centering
\resizebox{\textwidth}{!}{
\begin{tabular}{|l|lllll|l|r|r|}
\hline
\textbf{\begin{tabular}[c]{@{}l@{}}Reward\\ function\end{tabular}} & \textbf{Quality} & \textbf{Fluent} & \textbf{Diverse} & \textbf{Related} & \textbf{Empathy} & \textbf{Total}  & \textbf{Votes} & \textbf{\begin{tabular}[c]{@{}l@{}}Human\\ reward\end{tabular}}  \\ \hline
Conv. len.      & 2.20 $\pm$.40             & 3.61 $\pm$.53           & 3.02 $\pm$.52            & 2.25 $\pm$.46            & 2.48 $\pm$.45            & 13.57 $\pm$1.84          & -0.04          & -0.01  \\
Infersent Coher.   & 1.93 $\pm$.34          & 3.50 $\pm$.45           & 2.37 $\pm$.45            & 2.11 $\pm$.45            & 2.52 $\pm$.48            & 12.43 $\pm$1.75          & -0.02          & -0.01  \\
User laughter   & 1.96 $\pm$.38             & 3.56 $\pm$.48           & 2.33 $\pm$.51            & 1.93 $\pm$.42            & 3.20 $\pm$.55            & 12.98 $\pm$1.60          & -0.15          & -0.01 \\
User Word Len  & 2.11 $\pm$.32              & 3.96 $\pm$.44           & 3.04 $\pm$.45            & 2.04 $\pm$.35            & 2.55 $\pm$.46            & 13.70 $\pm$1.44          & 0.06           & \textbf{0.04}  \\
Manual votes    & 2.14 $\pm$.38             & 3.47 $\pm$.45           & 2.91 $\pm$.47            & 2.07 $\pm$.39            & 2.42 $\pm$.46            & 13.00 $\pm$1.65          & -0.03          & 0.01 \\
Sent. trans.    & 2.02 $\pm$.31             & 3.71 $\pm$.49           & 2.98 $\pm$.50            & 2.04 $\pm$.42            & 2.84 $\pm$.48            & 13.60 $\pm$1.63          & 0.03           & 0.01  \\
Bot Question        & 2.29 $\pm$.37         & \textbf{4.31 $\pm$.50}  & \textbf{3.31 $\pm$.52}   & 2.20 $\pm$.40            & 2.60 $\pm$.41            & 14.71 $\pm$1.63          & 0.06           & \textbf{0.04}  \\
User Sentiment       & \textbf{2.47 $\pm$.32}   & 4.05 $\pm$.45           & 3.23 $\pm$.46            & \textbf{2.42 $\pm$.39}            & \textbf{3.23 $\pm$.55}   & \textbf{15.40 $\pm$1.49} & \textbf{0.09}  & \textbf{0.04}  \\ \hline
\end{tabular}}
\caption{Interactive human evaluation of WOP trained with different reward functions on VHRED-EI model. Ratings are Likert scale with 95\% confidence interval ($n=45$), votes and human reward are $z$-scores.}
\label{tab:rewards_human_EI}
\end{table*}

\subsection{Computing Infrastructure}
Each RL model was trained on a NVIDIA GeForce GTX 1080 GPU. Training models for 2000 epochs took approximately 30 minutes for each model. The runtime for training the VHRED baseline models is around 6 hours. The speediness of training the RL models illustrates the scalability of RL training in improving dialog models for specific features. 

\subsection{Model Validation and Evaluation}
We use interactive human evaluation through an online chat interface. Human participants are recruited using Amazon Mechanical Turk and rate either 7 or 8 bots each. Participants were instructed to continue the conversation through at least 6 human responses. After the conversation, participants are asked to rate each bot in terms of \textit{Quality}, \textit{Fluency}, \textit{Diversity}, \textit{Contingency}, and \textit{Empathy} on a 7-point Likert scale. A detailed example of the chat and interaction platform can be found in Section \ref{sec:appendix-interactive}. Since our models are evaluated using interactive chat, we also validate our models through interactive chat and rate the models while tuning hyperparameters. The authors interacted with and rated bots during to validate bots.

\section{Offline-RL with VHRED with Emotion and Infersent Regularization}
We also conducted experiments using each offline RL algorithm with a Sentiment and Infersent regularized VHRED Model. As described in Section \ref{sec:appendix-model-details}, by adding about 20 million extra parameters to the VHRED model in order to better achieve semantic coherence and sentiment contingency, the VHRED-EI (Emotion and Infersent regularized) model is a better performing baseline in terms of human ratings \cite{ghandeharioun2019approximating}. 

We conducted the same human experiments where we recruited participants from Amazon Mechanical Turk to chat with and rate each dialog model. We found similar results as presented in our main paper. While our KL-control models achieved higher qualitative ratings than the other offline RL algorithms, none of the RL models received higher qualitative ratings than the VHRED-EI Model (Table \ref{tab:techniques_human_EI}). We also replicated training the KL-Control $\Psi$ model on single rewards and found that training on User Sentiment elicited the highest human qualitative ratings (Table \ref{tab:rewards_human_EI}). This consistent with our results on the VHRED model. 

\begin{figure*}[t]
  \centering
  \begin{subfigure}[t]{0.25\textwidth}
  \includegraphics[width=\textwidth]{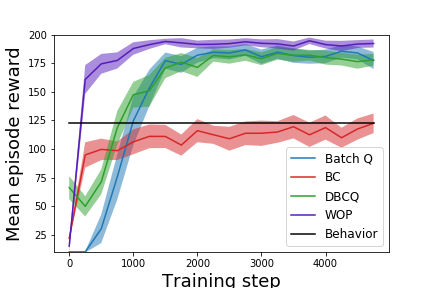}
  \caption{Full buffer}
\end{subfigure}
\begin{subfigure}[t]{0.24\textwidth}
  \includegraphics[width=\textwidth]{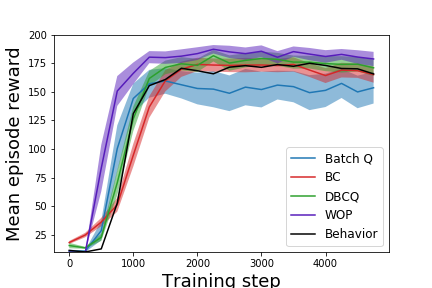}
  \caption{Concurrent}
\end{subfigure}
\begin{subfigure}[t]{0.24\textwidth}
  \includegraphics[width=\textwidth]{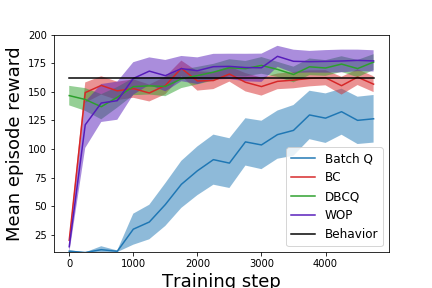}
  \caption{Expert demonstrator}
\end{subfigure}
\begin{subfigure}[t]{0.25\textwidth}
  \includegraphics[width=\textwidth]{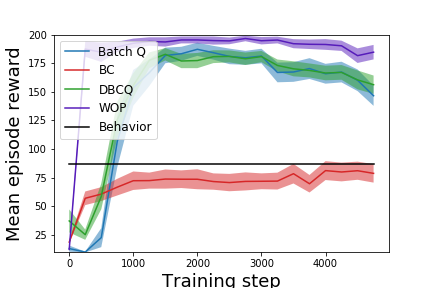}
  \caption{Noisy demonstrator}
\end{subfigure}
\caption{Comparison of batch RL algorithms in \textit{Cartpole-v0} for different offline learning conditions. WOP consistently exceeds the performance of Batch Q-learning, Behavioral Cloning (BC), DBCQ, and the Behavior policy used to generate the batch data. Error bars show 95\% \textit{CI} of the mean over 50 trials.}
  \label{fig:traditional_rl}
\end{figure*}

\begin{figure*}[t]
  \centering
  \begin{subfigure}[t]{0.25\textwidth}
  \includegraphics[width=\textwidth]{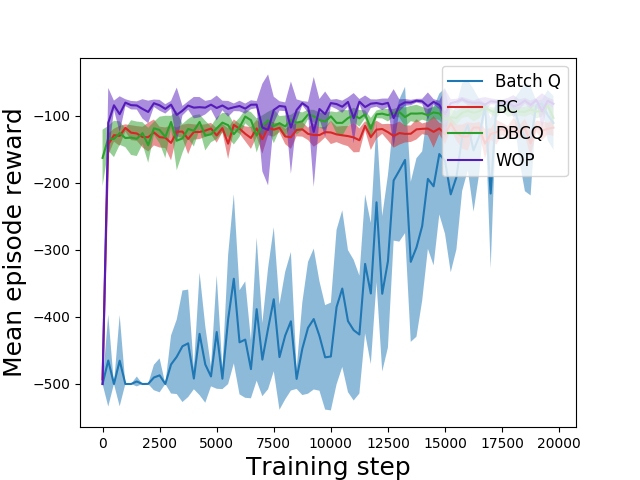}
  \caption{Full buffer}
\end{subfigure}
\begin{subfigure}[t]{0.24\textwidth}
  \includegraphics[width=\textwidth]{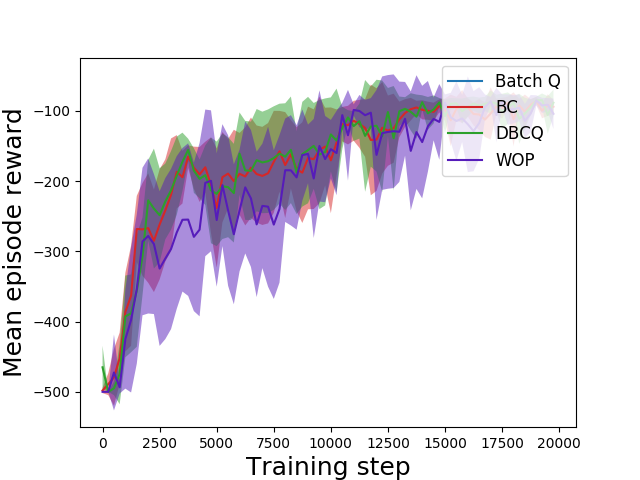}
  \caption{Concurrent}
\end{subfigure}
\begin{subfigure}[t]{0.24\textwidth}
  \includegraphics[width=\textwidth]{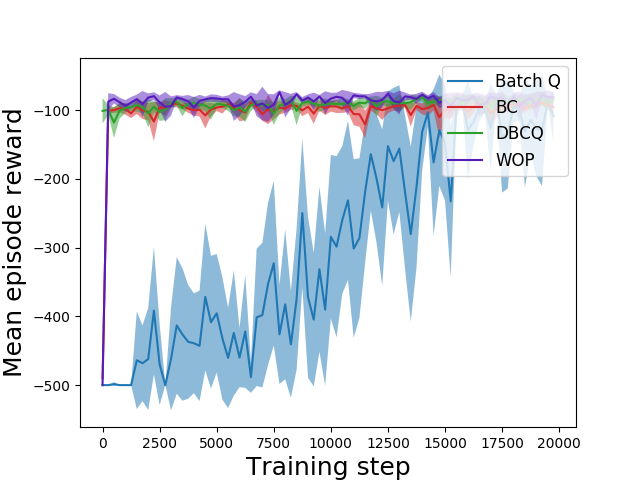}
  \caption{Expert demonstrator}
\end{subfigure}
\begin{subfigure}[t]{0.25\textwidth}
  \includegraphics[width=\textwidth]{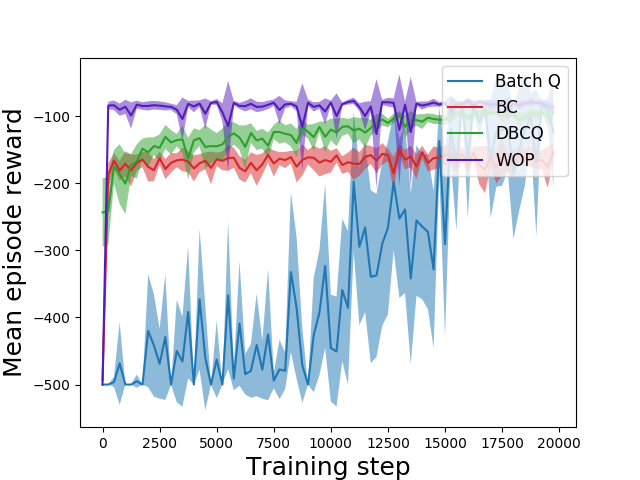}
  \caption{Noisy demonstrator}
\end{subfigure}
\caption{Comparison of batch RL algorithms for different offline learning conditions in \textit{Acrobot-v1}.}
  \label{fig:traditional_rl_acrobot}
\end{figure*}

\section{Traditional RL experiments}
\label{sec:tradrl}
To demonstrate the effectiveness of these techniques, we tested them on traditional RL tasks using the OpenAI gym \citep{openaigym}, focusing on the \textit{CartPole-v0} and \textit{Acrobot-v1} experiments. We first train an online $Q$-learning \textit{Behavior} policy, and store all $(s,a,r,s')$ experience samples into a replay buffer. We use this buffer to train a prior model of $p(a|s)$ using a Variational Auto-encoder. The VAE was trained to reconstruct the next state given the current state, $p(s'|s)$, using a mean-squared error loss.  The next action was predicted from the latent embedding $z$, meaning the model learned three functions: $z=f_e(s)$, $s'=f_d(z)$, and $a=f_a(z)$. For Cartpole, both the encoder and decoder were made up of two linear layers with 750 neurons each. The latent dimension of the VAE was size 256. For Acrobot, the encoder and decoder had only one layer of size 256 each, and the latent dimension was 64. 

This VAE is used as a part of both the DBCQ and WOP algorithms. We can also use it for imitation learning, by sampling actions directly from $p(a|s)$ to obtain Behavioral Cloning (BC). We benchmark all of these techniques against vanilla $Q$-learning on the batch data (\textit{Batch Q}). All $Q$-networks shared the same underlying architecture: three fully-connected layers of size [256, 128, 64], with ReLU activation between. All models were trained with the Adam optimizer \cite{kingma2014adam}. 

For each experiment, we ran 50 trials of each model with a different random seed each time.
The Behavior policy was trained for a total of 20,000 steps in the environment, so in the \textit{Full buffer} condition offline agents saw 20,000 experience samples. The Behavior policy typically converged before 10,000 steps, so in the \textit{Expert demonstrator} condition the offline agents received the last 10,000 experience samples from the trained agent. In the \textit{Concurrent} condition, offline agents saw a moving window of 1000 samples, since the online learner only used the most recent 1000 samples in the buffer for learning. The learning rate was .001, $\gamma=.99$, and $\epsilon$ decayed linearly from 1.0 to .01 over 2000 steps. 
The KL-constraint was computed as $D_{KL}[q(\tau)||p(\tau)] = \alpha\log p(a|s) - \beta\log \pi(a|s)$, where $\alpha=0.5$ and $\beta=0.1$. DBCQ sampled $n=2$ actions before selecting the best action based on the maximum $Q$-value; note that in this environment there are only 2 actions. For Cartpole we used the $\Psi$-learning loss, and for Acrobot we used the traditional $Q$-learning loss.

We experiment with four different conditions which vary the quality of the Behavior policy and the replay buffer data: a) \textit{Full buffer}: all experience samples experienced during online training are used for offline learning; b) \textit{Concurrent}: the offline learning algorithms see a sliding window of experience samples in the same order that the online learner experienced them; c) \textit{Expert demonstrator}: the buffer only contains experience generated by a fully trained online learner; and d) \textit{Noisy demonstrator}: the online learner has a high probability of acting randomly ($\epsilon=0.3$) and is thus a bad model of the optimal policy.

Figure \ref{fig:traditional_rl} shows the results. Across conditions, we see that WOP is able to outperform Batch $Q$, imitation learning (BC), DBCQ, and the original behavior policy. As expected, Imitation learning (BC) underperforms other techniques when the batch contains noisy or inexpert experience samples. However, when the batch contains only expert trajectories, Batch $Q$ fails to learn, because the batch does not cover the full state-action space well, increasing extrapolation error. DBCQ matches or outperforms BC and Batch $Q$ in all scenarios. However, because DBCQ acts by sampling from $p(a|s)$ as learned by the BC model, its performance suffers when the batch data is noisy or imperfect. In contrast, WOP is able to learn to trade-off staying close to the prior and obtaining higher reward, and consistently outperforms all other algorithms in this environment.  

\section{Additional results}
\begin{figure}[h]
  \begin{center}
   \includegraphics[width=.4\textwidth]{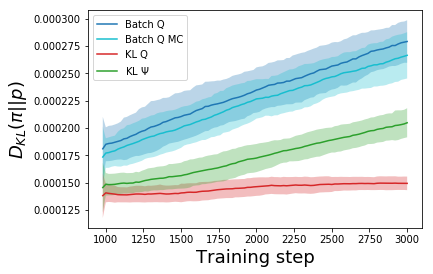}
  \caption{KL-divergence of the policy from the prior is lower with KL-control throughout training. Bands show $\sigma$.}
  \label{fig:kl}
  \end{center}
\end{figure} 

Figure \ref{fig:kl} shows the KL-divergence between RL policies and the prior language model throughout offline RL training. Without KL-regularization, the baseline RL models diverge quickly and continuously from the prior, losing information about realistic sequences. This figure also helps explain the poor performance of DBCQ in Table \ref{tab:techniques_human}. The underlying $Q$-network in DBCQ does not directly integrate the prior. As $Q$-learning causes the model to diverge from the prior, the $Q$-estimates of language generated according to the prior become unrealistic, and selects unrealistic actions. This results in highly `diverse' (random) generated utterances. Note that since we operate in discrete action space, we could not include the perturbation model originally proposed by \cite{fujimoto2018off}, which may be critical to achieving good performance with BCQ.

\section{Implicit Rewards Details}
\begin{figure*}[h]
  \centering
  \includegraphics[width=\textwidth]{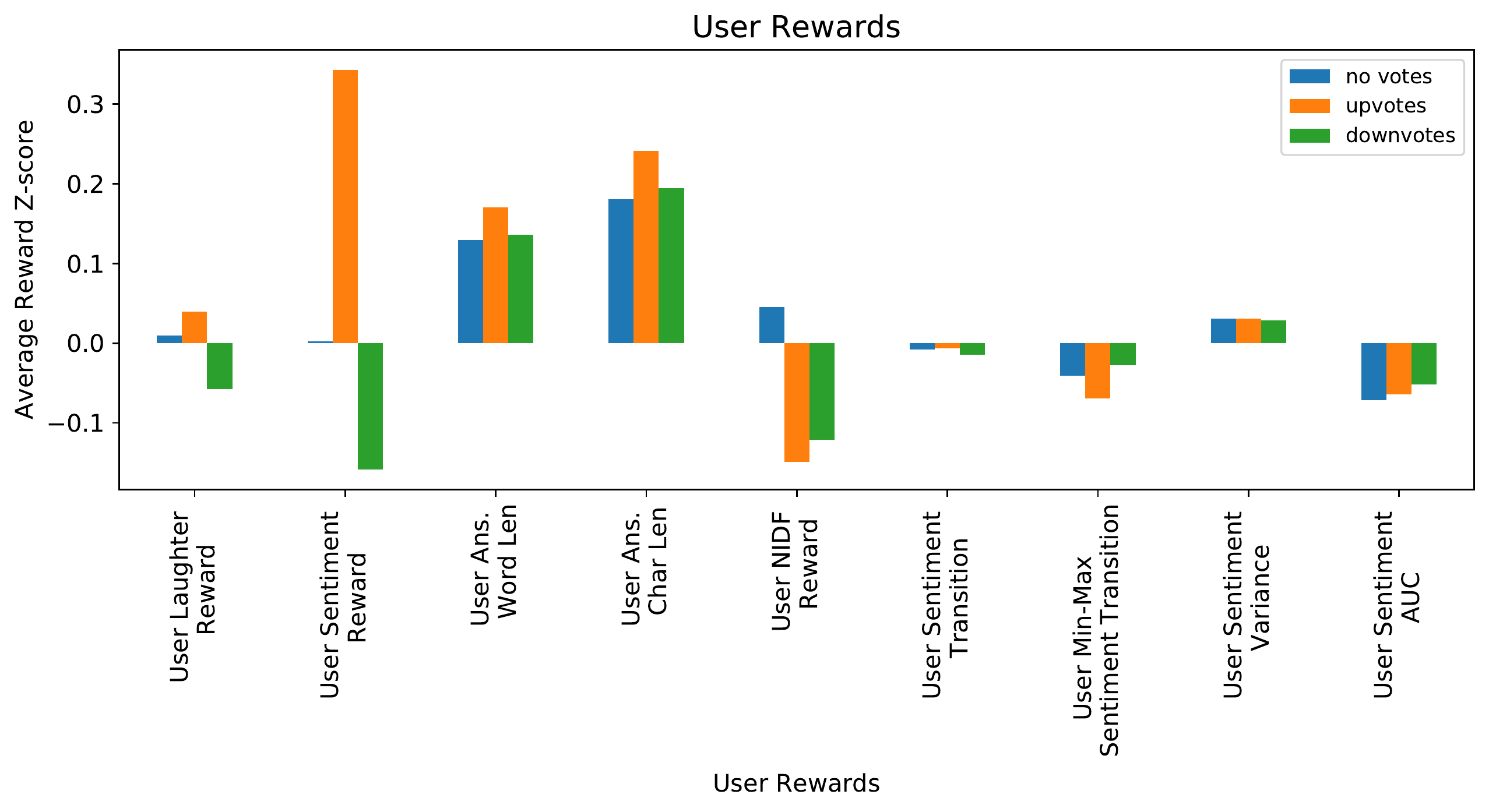}
  \caption{Mean z-scores for user-response-based rewards by manual vote}
  \label{fig:user-rewards}
\end{figure*}

\begin{figure*}[h]
  \centering
  \includegraphics[width=\textwidth]{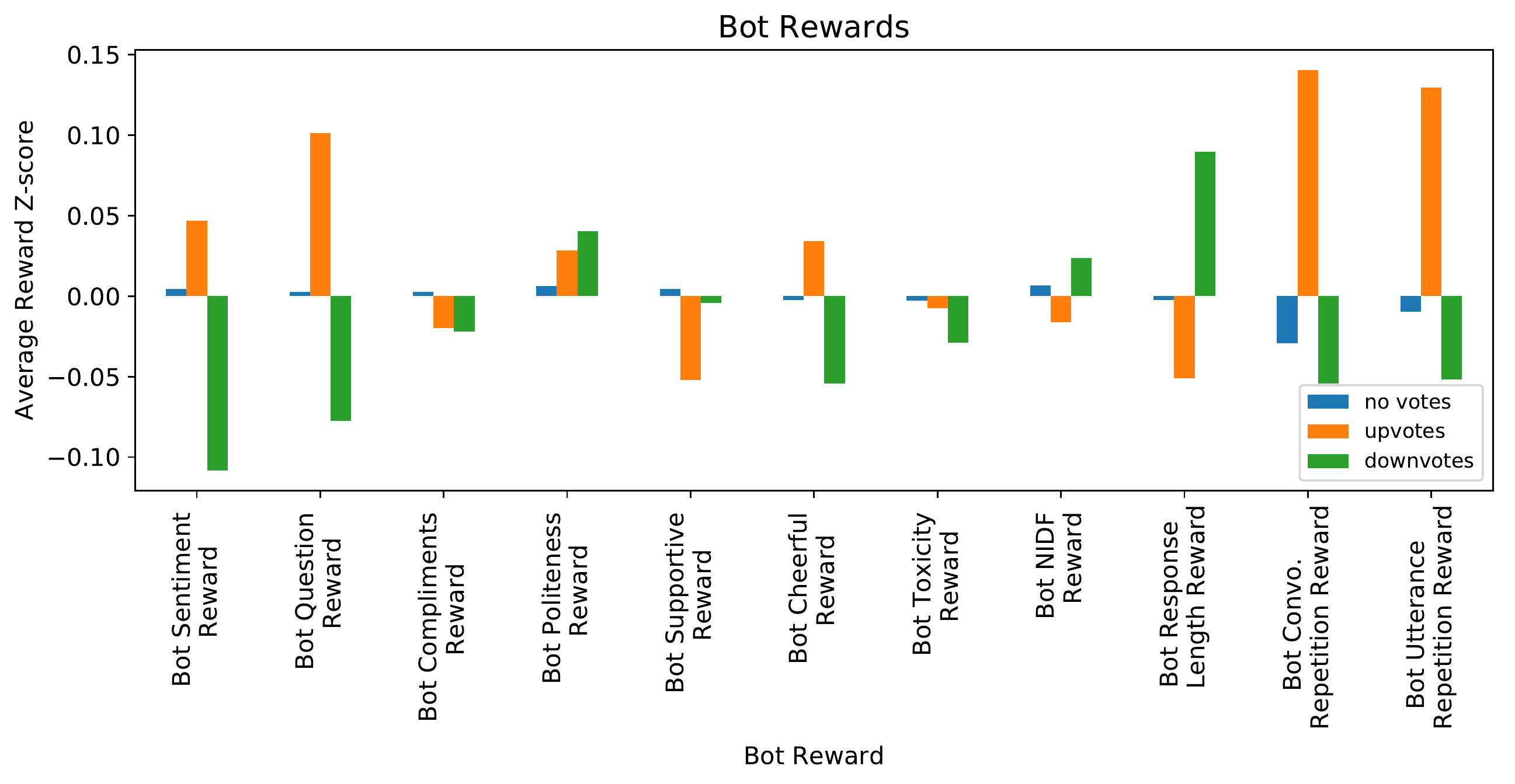}
  \caption{Mean z-scores for bot-based rewards by manual vote}
  \label{fig:bot-rewards}
\end{figure*}

\begin{figure*}[h]
  \centering
  \includegraphics[width=\textwidth]{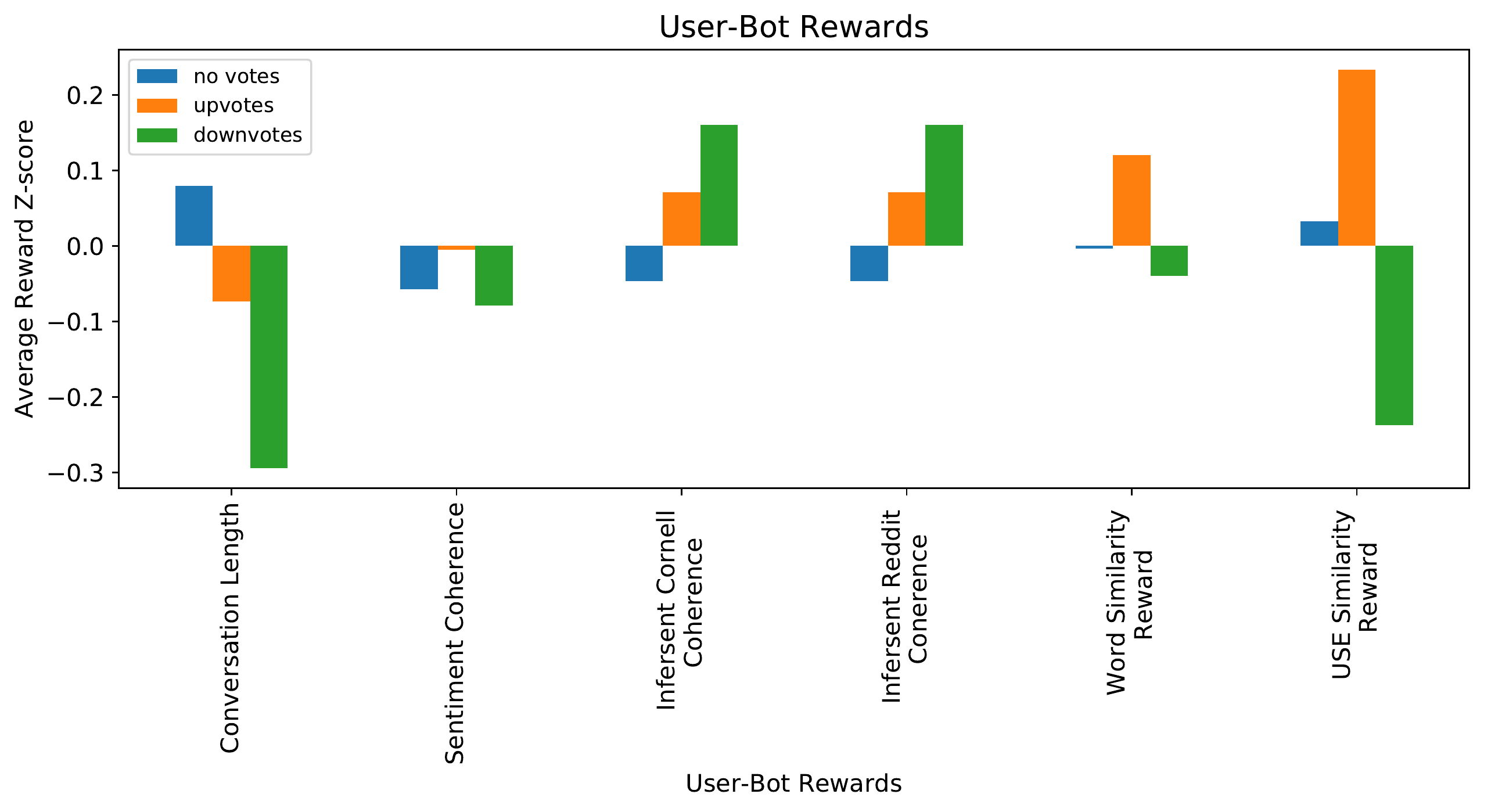}
  \caption{Mean z-scores for bot-user-based rewards by manual vote}
  \label{fig:user-bot-rewards}
\end{figure*}

The total reward used to train the bots is a combination of the rewards described in Table \ref{tab:reward_weights}. These rewards were selected based on the average z-score of rewards for utterances that were upvoted and downvoted. Figure \ref{fig:user-rewards} shows all the user rewards and that \textit{User Laughter} and \textit{User Sentiment} reward scores correlate with upvotes and downvotes. Figure \ref{fig:bot-rewards} shows all the bot rewards with \textit{Bot Sentiment}, \textit{Bot Laughter}, \textit{Bot Convo. Repetition}, and \textit{Bot Utterance Repetition} as rewards that correlate with manual votes. Figure \ref{fig:user-bot-rewards} shows the bot-user combined rewards, and that \textit{Word Similarity} and \textit{USE Similarity} are the rewards that correlate with manual up and downvotes. 
\begin{table}[h]
    \centering
    \begin{tabular}{|l|l|}
    \hline
        \textbf{Reward} & \textbf{Weight} \\ \hline
        User Sentiment & 0.10\\
        User Laughter & 0.10 \\
        USE Similarity & 0.15 \\
        Word Similarity & 0.15 \\
        Bot Question & 0.10 \\
        Bot Sentiment & 0.10 \\
        Bot Conversation Repetition & 0.15 \\
        Bot Utterance Repetition & 0.15 \\ \hline
    \end{tabular}
    \caption{Reward weights used for RL model training}
    \label{tab:reward_weights}
\end{table}

\subsection{Sentiment-based}
\begin{figure*}[h]
  \centering
  \begin{subfigure}[t]{0.5\textwidth}
  \includegraphics[width=\textwidth]{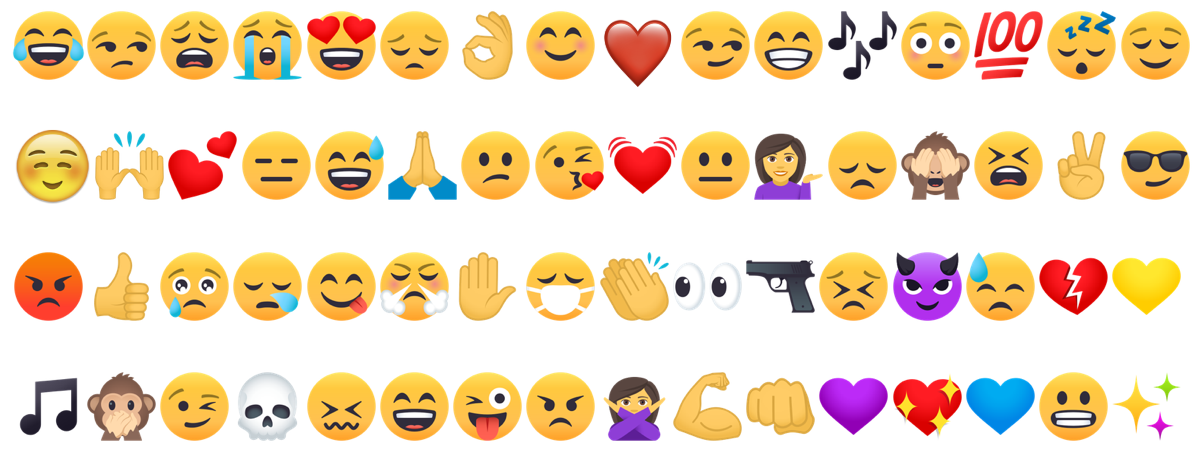}
  \caption{}
  \end{subfigure}
  \hspace{1cm}
  \begin{subfigure}[t]{0.35\textwidth}
  \includegraphics[width=\textwidth]{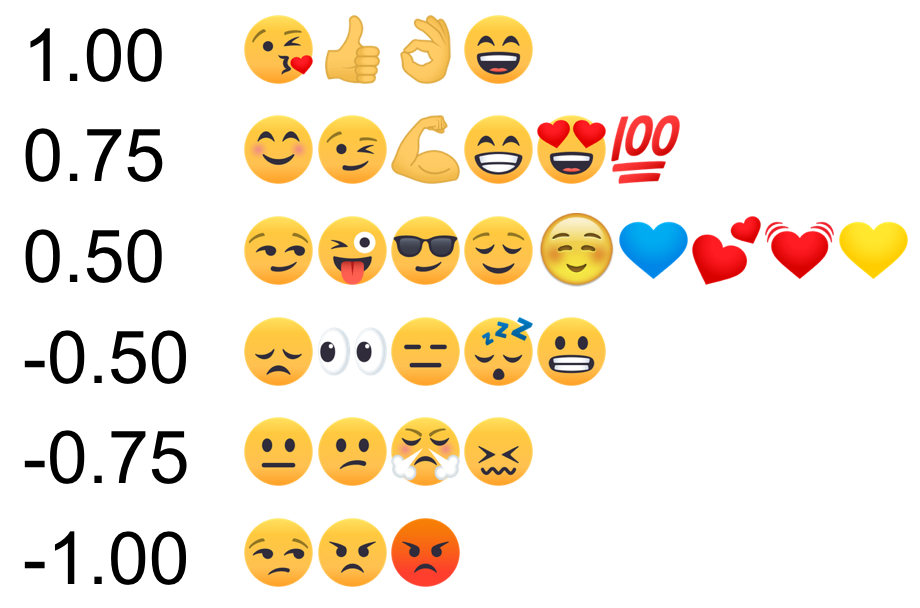}
  \caption{}
  \end{subfigure}
  \caption{(a) 64-most frequent emojis as predicted by \citep{felbo2017using} used for calculating emotion embeddings. (b) Assigned weights used in producing the sentiment reward from the predicted emoji values.}
  \label{fig:emojis}
\end{figure*}

To compute sentiment on short texts like conversation utterances, we leverage a state-of-the-art sentiment-detection model, which was trained on a massive amount of Twitter data to predict the emojis in tweets \citep{felbo2017using}. Transfer learning from this model to other tasks showed that it was able to significantly outperform a series of sentiment, irony, and sarcasm benchmarks. This DeepMoji model outputs a probability distribution over 64 most-frequently used emojis as shown in Figure \ref{fig:emojis}. After observing the performance of the model in detecting users' emotions in the domain of online chat, we define a set of weights over the emojis and calculate the weighted sum over an emotion embedding vector to derive a \textit{Sentiment} reward which is higher for positive sentiment and lower for negative sentiment. These weights are shown in Figure \ref{fig:emojis} (b). We also compute a sentiment-transition reward using the same score based on whether the peak positive sentiment occurred later in the conversation than the peak negative sentiment, reasoning that sentiment should improve over the course of the conversation. The \textit{Bot Sentiment} reward is the DeepMoji sentiment computed on the bot response, \textit{User Sentiment} reward is the value computed on the user response, and the \textit{Sentiment Coherence} reward is based on the similarly of user and bot sentiments.  

\subsection{Engagement-based}
Based on prior work \citep{zhou2018design}, we use the number of turns in the conversation as an indicator of the quality of the bot's performance. To distribute this reward over every utterance in the conversation, we take the total conversation length $N$, and compute the discounted reward for utterance $n < N$ as $\gamma^{N-n}N$ (\textit{Conversation Length}). We also reward each utterance with the number of words and characters in the user's response, which we refer to as \textit{User Ans. Word Len} and \textit{User Ans. Char Len}.  We also examine how long bot responses are with the \textit{Bot Response Length} reward. 

\subsection{Laughter}
Laughter has been shown to be very important to human affiliation \citep{provine1996laughter} and solidarity \citep{hay2000functions}. Therefore, we detect the number of occurrences of strings indicating laughter (e.g. `ha', `lol') in the user's response, and use this as a reward. Interestingly, we find that bots trained to maximize user laughter learn to be extremely supportive and cheerful compared to other bots (for definitions of supportive and cheerful see section \ref{sec:phrase-based}).

\subsection{Semantic similarity}
Language style matching has been shown to be a strong predictor of relationship initiation and stability \citep{ireland2011language}. While it would be ideal if our chatbots could intelligently adapt their conversation style to a new user, in reality most baseline dialog models struggle to maintain topic coherence, even over a few utterances (for an analysis of this effect, see \citep{ghandeharioun2019approximating}). Therefore we reward \textit{semantic similarity} between the user's input and the bot's response, to encourage the bot to stay on topic and produce reasonable answers. The \textit{Infersent Cornell Coherence} and \textit{Infersent Reddit Coherence} rewards are computed using a sentence embedding model trained on the Reddit and Cornell corpora respectively (described in section \ref{sec:appendix-model-details}). We use the Universal Sentence Encoder (\citep{conneau2017supervised}) to compute the \textit{USE Similarity} reward. We also directly compute word overlap as a reward as \textit{Word Similarity}. 

\subsection{Questions}
Asking questions is an important listening skill, and is linked to conversation management, attentiveness, and responsiveness \citep{bodie2012listening}. Therefore, we give the bot a reward of 0.5 if the utterance contains a question word (\textit{how, what, where, why, when, who}), and an additional 0.5 if it contains a question mark. We refer to this reward as \textit{Bot Question}. 

\subsection{Phrase based rewards}
\label{sec:phrase-based}
After training the bots on these rewards, we noticed a shift in the distribution of their language towards more polite, cheerful, and supportive speech. Therefore, we designed post-hoc metrics to measure these qualities, which are based on counting whether a subset of phrases is present in an utterance.

\textbf{Compliment phrases:} \textit{you are beautiful, you are so beautiful, you're beautiful, you're beautiful,
                   you are the best, you're the best, i like you, you're a good,
                   you re a good, i love the way you}

\textbf{Politeness phrases:} \textit{if I may; may I; please; thanks; no worries; if you don't mind; have a great day; I'm sorry}.

\textbf{Supportive phrases:} \textit{you're right; you are right; you're not alone; you are not alone; congrats; that's a good idea; that is a good idea; you'll be fine; you will be fine; you'll be okay; you will be okay; it will get better; sorry you're going through; sorry you are going through; if it makes you feel better; if it makes you feel any better; keep your head up; keep it up; I'm in a similar situation; I am in a similar situation; you'll get it; you will get it; happy for you; I'm in the same boat; I am in the same boat; if you feel like you need to vent}.             

\textbf{Cheerful phrases:} \textit{nice to hear; happy; excited; really nice; glad; the best; great; good time; looking forward; beautiful}.

\subsection{Toxicity}
We also want to discourage our bot from malicious or offensive language. \citet{saleh2019hierarchical} incorporate a Toxicity Classifier trained with data from the Toxic Comment Classification Challenge\footnote{\url{https://www.kaggle.com/c/jigsaw-toxic-comment-classification-challenge}} as a reward in the training hierarchical RL dialog models. We compute Toxicity reward scores using this classifier as \textit{Bot Toxicity} (e.g. lower toxicity score, higher \textit{Bot toxicity} reward). 

\subsection{Specificity}
Specificity within a conversation is valuable in avoid exchanging vacuous phrases back and forth. However building a chit-chat bot without a knowledge graph back-end limits the level of substance that can be incorporated into a conversation. We use the approach from \citep{see2019makes} of computing normalize IDF to create more specificity in the conversation. We compute NIDF on both user (\textit{User NIDF}) and bot (\textit{Bot NIDF}) text. 

\subsection{Repetition}
While minimizing repetition is a common implicit goal of dialog systems, we will explicitly optimize for reducing repetition through repetition rewards. We compute utterance repetition by the number of non-unique words in each utterance as \textit{Bot Utterance Repetition Reward}. We compute conversation repetition by the number of non-unique words in each conversation as \textit{Bot Convo. Repetition Reward}. These rewards are negated since we want a higher reward score for less repetition. We also remove stop words in the computation of non-unique words.

\section{Interactive bot platform details}
\label{sec:appendix-interactive}
\begin{figure*}[t]
  \centering
  \includegraphics[width=\textwidth]{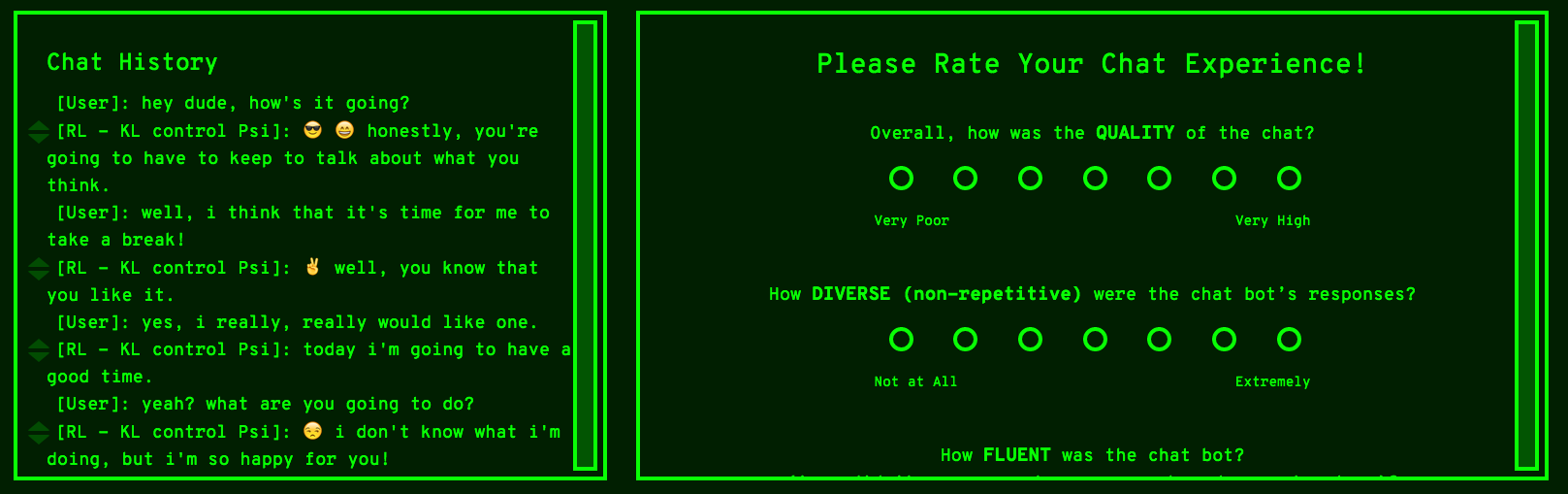}
  \caption{Interactive evaluation ratings page used to collect evaluations}
  \label{fig:interactive-rating}
\end{figure*}

To collect data from humans interacting with our bots, we built a platform for hosting deep neural network dialog models online on GPU for fast, real-time inference. Figure \ref{fig:interactive-rating} shows an example of the interface, in which users are able to rate the bots after talking to them for at least three turns. 

Note that during the chat, annotators can optionally click the up and down arrows beside each chatbot response to give feedback on the specific utterance. Once 6 or more turns of the conversation has taken place, participants may click ``Close Chat and Rate" to get to the rating screen. 

We train our RL models based on chat data collected on this platform. Currently, the conversations contain Personally Identifiable Information such as user name, age, location, etc. We obtained for IRB approval for this study and cannot release the conversations at this time in their current form.  

\subsection{Website server setup and configuration}
\label{sec:appendix-website}

The server was hosted on a Google Cloud Platform virtual instance with 64GB of RAM and a NVIDIA Tesla P100 graphics card. The backend was a Django program being served by NGINX and uWSGI. For simplicity, we opted to have the Django process import the chatbots into the same Python process as Django, rather than have the two connect to each other via other means such as sockets. This configuration decreased development time and increased reliability, but it would need to be revisited if the server needed to scale several orders of magnitude past what was required for this study. The current configuration was still able to support hundreds of simultaneous users and host more than 30 bots concurrently.

The chatbots were kept in a separate project from the Django project and maintained separately from the server code. Each chatbot extended an abstract class that defined key methods for the Django program to use, and was registered to a globally accessible dictionary via a decorator. The Django project was provided the path to the Chatbots project in its PYTHONPATH, so it could import the dictionary in which all the chatbot objects had been registered and use that to dynamically determine which chatbots were available and to access them in its views.

It is important to note that the chatbots used PyCUDA, and PyCUDA does not work in a multiprocessing environment. Because of this, uWSGI needed to be configured to only have one python process and to disable any attempt at multiprocessing. Furthermore, the chatbots required substantial startup times, so all chatbots are kept in memory at all times in the Django process. In order to keep all the chatbots in memory concurrently, we needed a very high amount of RAM on our server and opted for a 64GB virtual instance, and a GPU with 16GB RAM. This combination of CUDA to run the chatbots on the GPU with a high amount of RAM to keep all bots in memory at the same time resulted in incredibly fast server response times, with effectively no increase in response time when using the bots in requests compared to requests that did not.

For further information and instructions on server configuration, please read the server documentation available at \url{https://github.com/asmadotgh/neural_chat_web}. We hope that this platform will allow others to host their own bots and evaluate them in an interactive setting. 

\end{document}